\documentclass[10pt,twocolumn,letterpaper]{article}

\usepackage{cvpr}
\usepackage{times}
\usepackage{epsfig}
\usepackage{graphicx}
\usepackage{amsmath}
\usepackage{amssymb}
\usepackage{float}
\usepackage{footnote}
\usepackage{url}
\usepackage{arydshln}

\usepackage[breaklinks=true,bookmarks=false]{hyperref}

\usepackage{color}
\usepackage[normalem]{ulem}
\definecolor{darkgreen}{rgb}{0.0, 0.5, 0.0}

\def\SAdel#1{\bgroup\markoverwith{\textcolor{red}{\rule[0.5ex]{2pt}{1pt}}}\ULon{#1}}

\cvprfinalcopy 

\def\cvprPaperID{****} 

\pdfoutput=1

\let\svmaketitle\maketitle
\def\maketitle{\svmaketitle\thispagestyle{empty}}

\begin{document}

\title{Ensemble based discriminative models for Visual Dialog Challenge 2018}

\author{Shubham Agarwal  \thanks{Adeptmind Scholar} \\
Heriot Watt University \\
Edinburgh, UK \\
{\tt\small sa201@hw.ac.uk}
\and
Raghav Goyal \thanks{work done at TwentyBN}\\
University of British Columbia\\
Vancouver, Canada\\
{\tt\small rgoyal14@cs.ubc.ca}
}

\maketitle

\begin{abstract}
This manuscript describes our approach for the Visual Dialog Challenge 2018. We use an ensemble of three discriminative models with different encoders and decoders for our final submission. Our best performing model on `test-std' split achieves the NDCG score of $55.46$ and the MRR value of $63.77$, securing third position in the challenge.
\end{abstract}

\section{Introduction}
 
Visual dialog \cite{visdial} is an interesting new task combining the research efforts from Computer Vision, Natural Language Processing and Information Retrieval. While \cite{teney2018tips} presents some tips and tricks for VQA 2.0 Challenge, we follow their guidelines for the Visual Dialog challenge 2018. Our models use attention similar to \cite{anderson2018bottom} to get object level image representations from Faster R-CNN model \cite{ren2015faster}. We experiment with different encoder mechanisms to get representations of conversational history. 
\section{Models}
\label{sect:models}
Common to all the models, we initialize our embedding matrix with pre-trained Glove word vectors of 300 dimensions using 6B tokens \footnote{https://nlp.stanford.edu/projects/glove/}. Out of $11319$ tokens present in the dataset, we found $188$ tokens missing from the pre-trained Glove embeddings, so we manually map these tokens to words conveying semantically similar meaning, e.g. we map over ten variations of the word ``yes'' - misspelled or not picked up by tokenizer - ``*yes", ``yesa", ``yess", ``ytes", ``yes-", ``yes3", ``yyes", ``yees", etc.

For image features, we extract Faster R-CNN features with ResNet-101 backbone trained on Visual genome \cite{krishna2017visual} dataset, similar to \cite{anderson2018bottom}. We use an adaptive number of object proposals per-image ranging from $10$ to $100$ generated using a fixed confidence threshold and each object is then associated with $2048$-dimensional mean-pooled features using ROI pooling. We use discriminative decoding throughout our models.

We first describe our models individually and then the ensembling technique that we employ. In the following, \textit{MN} denotes Memory Networks to encode conversational history, \textit{RCNN} signify R-CNN for object level representations of an image, \textit{Wt} represents additional linear layer in the decoder, and \textit{LF} a late fusion mechanism as defined in \cite{visdial}.

\begin{figure*}
\centering
\resizebox{\textwidth}{!}{
\begin{tabular}{c}
\includegraphics[scale=0.75]{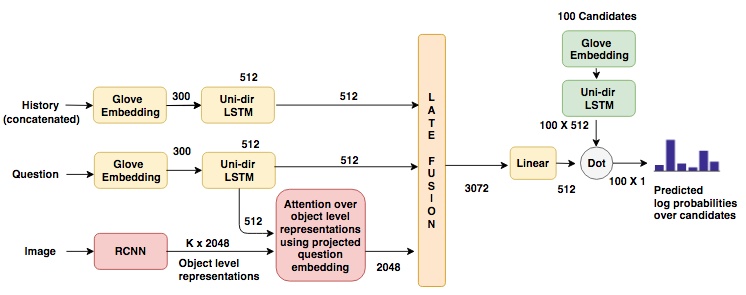}
\\
\caption{LF-RCNN model as defined in Section \ref{sect:models}. Number indicates the output dimension of each layer.}
\label{fig:lf}
\\
\includegraphics[scale=0.75]{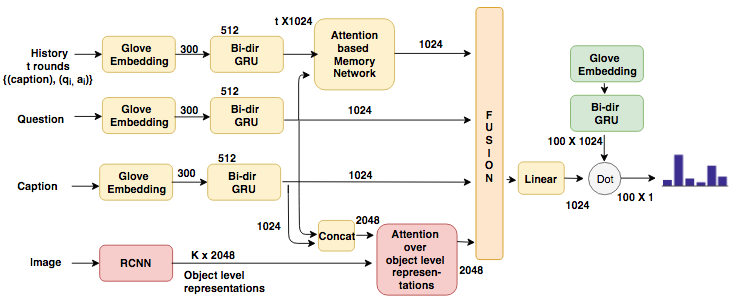}
\\
\caption{MN-RCNN model}
\label{fig:mn-rcnn}
\\
\includegraphics[scale=0.75]{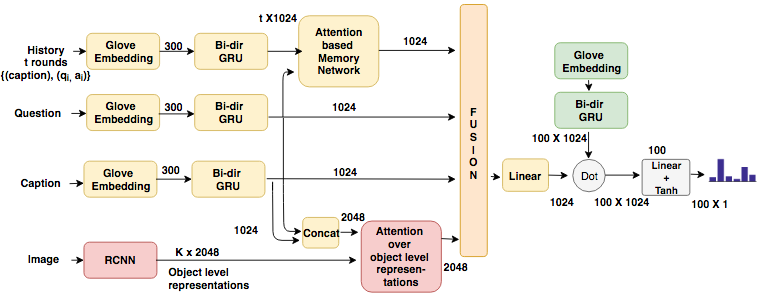}
\\
\caption{MN-RCNN-Wt model}
\label{fig:mn-rcnn-wt}
\end{tabular}}
\end{figure*}

\paragraph{LF-RCNN}
\textit{Late fusion encoder \cite{visdial} with concatenated history.} We use two-layered LSTMs with $512$ hidden units for embedding questions and history. The object-level features are weighed using only question embeddings. The word embeddings from Glove vectors are frozen and are not fine-tuned. Figure \ref{fig:lf} gives an overview of the architecture.

\paragraph{MN-RCNN}
\textit{Memory network encoder \cite{visdial} with bi-directional GRUs and word embeddings fine-tuned.} Object-level features are weighed by question and caption embedding. The rest of the scheme is same as above. (Figure \ref{fig:mn-rcnn})

\paragraph{MN-RCNN-Wt}
Same as above but with an additional linear layer applied to the dot product of candidate answer and encoder output, and gated using \textit{tanh} function. Compare Figure \ref{fig:mn-rcnn-wt} with Figure \ref{fig:mn-rcnn} 

\begin{figure*}[ht]
\centering
\includegraphics[scale=0.595]{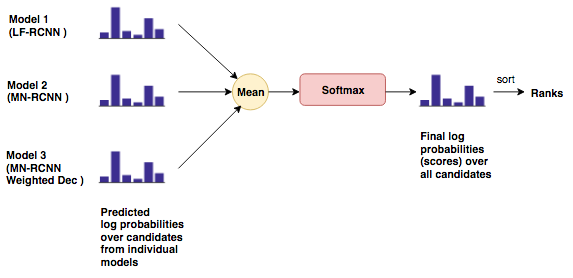}
\caption{Ensemble using mean of log probabilities from the three individual models.}
\label{fig:ensemble}
\end{figure*}

\paragraph{Ensembling}
We ensembled final layer's log-softmax output - which is a distribution over candidate answers for each round (Figure \ref{fig:ensemble}). We use the three models described above and take the mean of the results (we also tried taking maximum of the results but found mean to perform better). We also tried ensembling a subset of the above three models, but found the combination of all three to outperform the rest.

\begin{table*}[ht]
\centering
\begin{tabular}{lcccccc}
\hline
\textbf{Model} & \textbf{NDCG (x 100)} & \textbf{MRR (x 100)} & \textbf{R@1}  & \textbf{R@5}   & \textbf{R@10}  & \textbf{Mean} \\ \hline \hline
LF-RCNN & 51.69 & 61.03 & 47.03 & 77.83 & 87.55 & 4.70
\\ 
MN-RCNN & 53.59 & 61.25 & 46.78 & 79.43 & 87.93 & 4.63
\\ 
MN-RCNN-Wt & 53.20 & 61.50 & 47.10 & 78.7 & 88.38 & 4.54
\\ 
\hline
\textbf{Ensemble (all three)} & \textbf{55.46} & \textbf{63.77} & \textbf{49.8} & \textbf{81.22} & \textbf{90.03} & \textbf{4.11} 
\\ 
\hline
\end{tabular}
\caption{Results for the challenge on test-std. We took an ensemble of best performing models for the final submission.}
\label{table:results-test}
\end{table*}

\begin{table}[ht]
\centering
\resizebox{0.48\textwidth}{!}{
\begin{tabular}{lccccc}
\hline
\textbf{Model} & \textbf{MRR} & \textbf{R@1} & \textbf{R@5} & \textbf{R@10} & \textbf{Mean}  \\
\hline \hline
Baseline & 57.57 & 42.98 & 74.64 & 84.91 & 5.48 \\
MN & 59.24  & 44.64 & 76.48 & 86.41 & 5.14 \\
MN-Wt & 59.54 & 44.98 & 77.10 & 86.38 & 5.03  \\
\hdashline[0.5pt/2pt]
LF-RCNN & 61.94  & 48.08 & 79.04 & 88.23 & 4.61 \\
MN-RCNN & 62.99  & 49.07 & 80.13 & 88.74 & 4.45 \\
MN-RCNN-Wt & 63.11  & 49.29 & 80.10 & 89.09 & 4.43 \\
\hline
\end{tabular}}
\caption{Results on validation set. We show the impact of using additional gated linear layer in decoder. Compare MN with MN-Wt.}
\label{table:results-val}
\end{table}

\section{Experiments and Results}

We used Pytorch\footnote{\url{https://pytorch.org/}} \cite{paszke2017automatic} for implementation.\footnote{We are thankful to the organizers for providing a starter code in Pytorch on which we build our model. \url{https://github.com/batra-mlp-lab/visdial-challenge-starter-pytorch}} In our experiments, we find that fine-tuning initialized Glove embeddings performed better than frozen embeddings. Object level representations play a critical role to generate a correct response from the model. Eventually, we use an ensemble of all the models described above for our final submission. Table \ref{table:results-val} summarizes our results on validation set while Table \ref{table:results-test} on \textit{Test-Standard} split.

\section{Conclusion}

We experimented with discriminative models for our submission. Object level image representations gave a huge uplift in the evaluation metrics. Bi-directional GRUs constantly performed better than uni-directional LSTMs with Memory Networks outperforming Late fusion encoders for encoding conversational history. We even found that fine-tuning Glove embeddings performed better than their counterparts. Our final submission is an ensemble of three discriminative models and achieve the NDCG of 55.46 on test-std.

{\small
\bibliographystyle{ieee}
\bibliography{egbib}
}

\end{document}